\documentclass[accepted]{uai2024} 
                        
\usepackage[american]{babel}

\usepackage{natbib} 
\usepackage{mathtools} 
\usepackage{booktabs} 
\usepackage{tikz} 

\usepackage{enumitem}
\usepackage{xcolor}
\usepackage{amsmath}
\usepackage{amssymb}
\usepackage{bm}
\usepackage{bbm}
\usepackage{multirow}

\usepackage{microtype}
\usepackage{graphicx}
\usepackage{subfigure}
\usepackage{booktabs} 

\newcommand{\err}{\mathrm{err}}

\newcommand{\Appendix}[1]{the full version for}

\newtheorem{definition}{Definition}

\newcommand{\R}{\mathbb{R}}

\newcommand{\cD}{\mathcal{D}}

\newcommand{\cH}{\mathcal{H}}

\newcommand{\cN}{\mathcal{N}}

\newcommand{\cY}{\mathcal{Y}}
\newcommand{\cX}{\mathcal{X}}

\definecolor{colorY}{rgb}{0.7 , 0.7 , 0.2}

\title{Bayesian Neural Networks with Domain Knowledge Priors}

\author[1]{Dylan Sam$^*$}
\author[1]{Rattana Pukdee$^*$}
\author[1]{Daniel P. Jeong}
\author[1]{Yewon Byun}
\author[1,2]{J. Zico Kolter}

\affil[1]{%
    Carnegie Mellon University\\
}
\affil[2]{%
    Bosch Center for AI
}
  
\begin{document}
\maketitle

\def\thefootnote{*}\footnotetext{Equal Contribution.}

\begin{abstract}

Bayesian neural networks (BNNs) have recently gained popularity due to their ability to quantify model uncertainty. However, specifying a prior for BNNs that captures relevant domain knowledge is often extremely challenging. In this work, we propose a framework for integrating general forms of domain knowledge (i.e., any knowledge that can be represented by a loss function) into a BNN prior through variational inference, while enabling computationally efficient posterior inference and sampling. 
Specifically, our approach results in a prior over neural network weights that assigns high probability mass to models that better align with our domain knowledge, leading to posterior samples that also exhibit this behavior.
We show that BNNs using our proposed domain knowledge priors outperform those with standard priors (e.g., isotropic Gaussian, Gaussian process), successfully incorporating diverse types of prior information such as fairness, physics rules, and healthcare knowledge and achieving better predictive performance. We also present techniques for transferring the learned priors across different model architectures, demonstrating their broad utility across various settings.

\end{abstract}

\section{INTRODUCTION}\label{sec:intro}

While recent advances in deep learning have led to strong empirical performance in many real-world settings, applying deep learning models to high-stakes domains (e.g., healthcare, criminal justice) necessitates correctness, as incorrect predictions can lead to serious consequences. At the same time, it is also crucial for deep learning models to faithfully represent their uncertainty in prediction and avoid making incorrect predictions with high confidence, especially in such domains. Unfortunately, deep learning models trained via empirical risk minimization have been shown to often make errors with high confidence, especially on data points that differ from those observed in the training data distribution \citep{hendrycks2016baseline, hendrycks2021many}. 
Moreover, these models often inherit undesirable biases present in their training data \citep{compas,obermeyer}, motivating the development of an approach for incorporating prior knowledge that can counteract and mitigate these issues.

\begin{figure*}[t]
    \centering
    \includegraphics[width=0.75\textwidth]{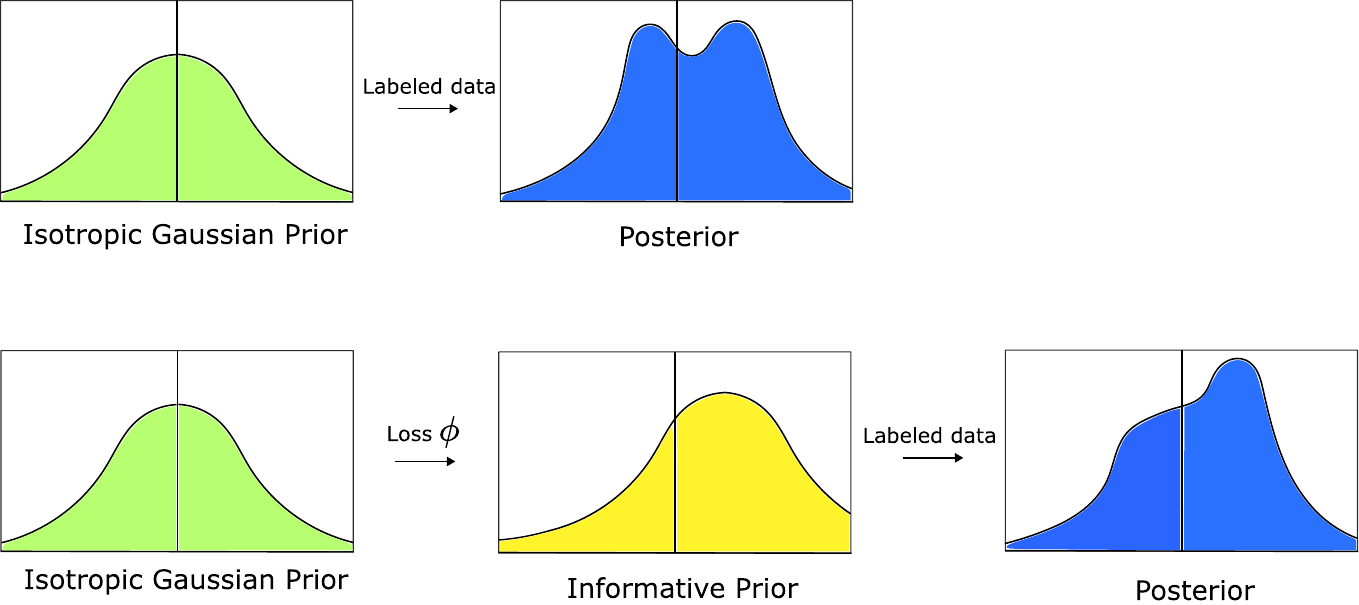}
    \caption{ Our framework (Banana; bottom row) compared to standard practice (top row) for training BNNs. Our method incorporates domain knowledge via a loss function $\phi$ to learn an intermediate step of an informative prior via a variational objective. The informative prior helps encourage models that exhibit desirable behavior. }
    \label{banana_fig}
    \vspace{-4mm}
\end{figure*}

One principled approach to achieving both correctness and a faithful representation of predictive uncertainty is with Bayesian neural networks \citep[BNNs;][]{mackay-thesis,mackay-1995,bnn-neal,bdl}.
In BNNs, we treat the neural network weights as random variables and specify a prior distribution over them to represent our a priori belief about what regions of the weight space are plausible. By Bayes' rule, we can then derive the posterior distribution over the weights conditioned on the training data. The posterior distribution allows us to account for the uncertainty in the model when making predictions, which is especially valuable when making high-stakes decisions.

In the Bayesian setting, selecting a good prior is crucial, and its misspecification for BNNs can force the posterior distribution to contract to suboptimal regions of the weight space \citep{prior-misspecification,gelman-2017,gelman2020bayesian,fortuin-priors}, which translates to suboptimal posterior predictive performance. Ideally, the prior should accurately reflect what relevant domain knowledge (e.g., physics rules) specifies as plausible functions for a given prediction problem and help mitigate any undesirable biases learned from the training data. However, the high-dimensionality of the weight space and the nontrivial connection between the weight and function spaces make specifying a prior that reflects domain knowledge extremely challenging \citep{nalisnick-2018,fortuin-priors}. Due to such difficulties, uninformative priors that enable tractable sampling and approximate inference are typically used in practice. The most widely used uninformative weight-space prior is the isotropic Gaussian prior \citep{hernandez-2015}. Recent works propose to directly specify a function-space prior (e.g., via Gaussian processes (GPs)) to encode functional properties such as smoothness and periodicity \citep{sun2018functional}. However, existing forms of informative priors are not flexible enough to represent broader forms of domain knowledge.

In this work, we propose a novel approach for incorporating much more \textit{general} forms of domain knowledge into BNN priors in a semi-supervised learning setup. The key challenge lies in both how to formulate and incorporate such knowledge into the prior and to ensure that this prior leads to computationally tractable posterior inference and sampling. In particular, we focus on domain knowledge for which we can formulate a 
\textit{loss function} $\phi$, such that it captures how well a particular model aligns with the given knowledge. We show that various forms of domain knowledge can be represented in this form (Section \ref{sec:data}).
For example, for a physics rule, we can define the loss function to measure how much a model's prediction of the state of a physical system violates the law of conservation of energy. 
As another example, if we want a vision model to ignore the background of an image, we can define the loss to be the norm of the gradient of the model's prediction with respect to the background pixels of the image.
To obtain an informative prior that incorporates such domain knowledge, we propose a variational inference approach to learn a low-rank Gaussian distribution that puts higher probability mass on model weights with low values of the loss $\phi$ on unlabeled data.
The low-rank Gaussian structure of the informative prior enables computationally efficient posterior inference.
We emphasize that with existing approaches for specifying informative priors, it is not clear how to incorporate similar forms of prior knowledge.

We demonstrate that using our learned informative priors for posterior inference in BNNs not only ensures better alignment with domain knowledge (i.e., lower values of $\phi$) but also improves predictive performance across many datasets, where various forms of domain knowledge (e.g., feature importance, clinical rules, fairness constraints) are available.
Notably, our approach outperforms BNNs that use an uninformative isotropic Gaussian prior, as well as those with more specialized---yet unable to flexibly incorporate general forms of domain knowledge---priors. 

We also present various techniques, based on maximum mean discrepancy \citep{gretton2012kernel} and moment matching (with SWAG \citep{maddox2019simple}), for \textit{transferring} a learned domain knowledge prior across different model architectures to increase their overall utility. In general, a BNN prior is \textit{architecture-specific}, i.e., we cannot directly use a prior learned for one BNN in another (e.g., with a different number of hidden layers or units). 
While relearning a new prior every time is one option, such an approach can be expensive and even infeasible in scenarios when we \textit{no longer have access to the loss} $\phi$. For example, clinical rules derived from patient data in one hospital\footnote{See \textbf{Thresholds Used for Defining $\phi_{\text{clinical}}$} in Appendix \ref{appendix-healthcare}.} may not be accessible in another due to privacy. 
Our empirical results demonstrate that we can efficiently transfer our priors
to different model classes, where models sampled from a transferred prior achieve significantly lower values of $\phi$ compared to models drawn from an isotropic Gaussian prior. 

\section{RELATED WORK}\label{sec:related-work}

\paragraph{Learning with Domain Knowledge.} 
Many researchers have focused on incorporating domain knowledge or explanations into increasingly black-box deep learning models. 
Some approaches directly regularize models to incorporate instances of such domain knowledge \citep{ross2017right, rieger2020interpretations, ismail2021improving}. 
For example, \citet{rieger2020interpretations} discourage models from using spurious patches in images for skin cancer detection tasks by penalizing models that place high feature importance on those patches. However, prior knowledge can also come in various forms beyond explanations, including rules from physics \citep{NEURIPS2018_842424a1, seo2021controlling}, weak supervision \citep{sam2022losses}, invariance \citep{simclr}, explicit output constraints for particular regions of the input space \citep{output-bnn-2020} or desirable properties such as fairness
\citep{zafar2017fairness, dwork2012fairness}. 
These works suggest that incorporating domain knowledge can lead to models that are more robust and perform better out-of-distribution. Existing work theoretically analyzes such simple incorporation of domain knowledge as constraints to show benefits in sample complexity \citep{pukdee2023learning}. 

In the Bayesian setting, existing works have studied directly regularizing posterior samples \citep{zhu2014bayesian, huang2023posterior}, but none have extensively studied how to obtain BNN priors that incorporate forms of domain knowledge as broad as those aforementioned.
Moreover, using informative priors is more computationally efficient than posterior regularization, as we only need to compute $\phi$ during the pretraining phase for the prior (see Section \ref{sec: learning informative priors}) and not for every posterior sample. Our method thus scales better when sampling a large number of posterior samples, allowing a more accurate approximation of the model posterior average. 

\paragraph{Priors in Bayesian Neural Networks.} 

As previously mentioned, the high-dimensionality of the weight space for BNNs makes specifying a prior that reflects aforementioned forms of domain knowledge extremely challenging \citep{nalisnick-2018,fortuin-priors}. 
Prior works propose to encode functional properties such as smoothness and periodicity by using a function-space prior (e.g., via GPs \citep{gpforml}) \citep{pmlr-v54-sun17b, sun2018functional,noise-contrastive-priors, tran2022all},  
to encode output constraints for particular regions of the input space into a weight-space prior \citep{output-bnn-2020}, or to use a set of reference models (e.g., simpler linear models) as priors to regularize the predictive complexity of BNNs \citep{pred-complexity-priors}. Our work differs from prior work in that we address more general notions of domain knowledge, such as feature importance and fairness, which are difficult to achieve with existing methods (e.g., how does one encode notions of fairness into a GP kernel?).

More recent works propose to leverage advances in self-supervised learning \citep{henaff-2020,simclr} to learn more informative and expressive priors from auxiliary, unlabeled data. \citet{sharma2023incorporating} propose to learn an informative prior by fixing the parameters of the base encoder to the approximate maximum a posteriori (MAP) estimate from contrastive learning \citep{simclr}. \citet{shwartz2022pre} propose to use a temperature-scaled posterior from a source task as a pretrained, informative prior for the target task, and empirically demonstrate that a BNN with an informative prior consistently outperforms BNNs with uninformative priors (e.g., isotropic Gaussian) and non-Bayesian neural network ensembles in predictive accuracy, uncertainty estimation, and data efficiency.  

We remark that, to the best of our knowledge, there are no other existing BNN methods that allow incorporating general forms of domain knowledge into BNN priors. The most relevant prior work by \citet{output-bnn-2020}, which encodes information by upweighting models that satisfy particular constraints on their output space, can be seen as a specific instance of our framework but is not easily applicable to most tasks considered in this paper.    

\section{PRELIMINARIES}

We consider a standard semi-supervised learning setting. Let $\cX$ be an instance space and $\cY$ be a label space. Let $\cD$ be a distribution over $\cX\times \cY$. We observe a training dataset of examples $X = \{(x_1, y_1), ..., (x_n, y_n) \}$ and unlabeled examples $X' = \{x_1', ..., x_k'\}$ drawn from $\cD$ and a marginal distribution $\cD_\cX$, respectively. We consider a class of neural networks $\mathcal{H} = \{h_w | h_w: \cX \to \cY\}$, which have weights $w$. Our goal is to learn a neural network $h$ (or a distribution over possible neural networks with a mean) that achieves the lowest loss, or
\begin{align*}
    \err(h): = \Pr_{(x,y) \sim \cD}(\ell(h_w(x), y)),
\end{align*}
where $\ell$ is the 0-1 loss for classification, and the $\ell_1$ or $\ell_2$ loss for regression.

One approach to capture model uncertainty is via BNNs, which models a distribution over neural networks via a distribution over \textit{weights}, $q(w)$. In practice, it is common to assume a standard isotropic Gaussian prior $q(w) = \prod_{i} \cN(w_i; 0, \sigma_i^2)$. This does not capture any prior knowledge about downstream tasks but is primarily used for its computational tractability. 
Given a prior $q(w)$ over neural network weights $w$ and labeled data $X$, we can sample from the posterior distribution using stochastic gradient Markov chain Monte Carlo methods such as Stochastic Gradient Hamiltonian Monte Carlo (SGHMC) \citep{chen2014stochastic} and Stochastic Gradient Langevin Dynamics (SGLD) \citep{welling2011bayesian}. 
In this work, we mainly use SGLD in our experiments (Sections \ref{sec:data}--\ref{sec:results}) but also consider MultiSWAG \citep{bdl} in our ablations (Section \ref{sec:ablation}).

\section{Domain Knowledge Priors for Bayesian Neural Networks}
While existing methods tackle specific desirable properties of a network (e.g., smoothness), it is unclear how to incorporate very general forms of domain knowledge into BNNs, as discussed in Sections \ref{sec:intro}--\ref{sec:related-work}.
We propose to achieve this by incorporating such information into a data-driven prior. 

\subsection{Domain Knowledge Loss}
 
First, we define our notion of domain knowledge. We propose to represent this as a loss function that measures the alignment of a particular model to our domain knowledge.

\begin{definition}
(Domain Knowledge Loss)
A domain knowledge loss function can be expressed as $\phi: \cH \times \cX \to \R$, which takes inputs $h \in \cH, x \in \cX$ and has $\phi(h, x) \geq 0.$ 
\end{definition}

We capture how well $h$ satisfies our domain knowledge at a point $x$ through this loss function, where a lower loss value implies that $h$ better satisfies the domain knowledge. This definition is quite general, and it is possible to define the loss $\phi$ to capture various notions of domain knowledge including physical rules and information about spurious correlations (see examples of these losses in Section \ref{sec:data}). We remark that these notions of domain knowledge are functions of the random input data $x$, and thus are difficult to directly encode in function space or via a kernel in a GP prior.

Given this definition of domain knowledge, we want our models to achieve low values of this loss, say having $\mathbb{E}_{x \sim \cD_\cX}[\phi(h, x)] \leq \tau$, where $\tau$ is some threshold. We also remark that this loss can be evaluated solely on unlabeled data, which yields nicely to using this for pretraining or learning priors. Considering losses that use information about labels could be potentially interesting, especially in the case of certain fairness metrics, e.g., equal odds and disparate impact \citep{hardt2016equality, mehrabi2021survey}. 

In the frequentist setting, we can incorporate such domain knowledge by simply adding a regularization term based on this surrogate loss \citep{ross2017right, rieger2020interpretations, pukdee2023learning}. For a loss function $\ell$, this yields a regularized objective given by
\begin{align}
    \min_{h \in \cH} \frac{1}{n}\sum_{i=1}^n \ell(h, x_i,y_i) + \lambda \cdot \frac{1}{k}\sum_{i=1}^k \phi(h, x'_i),
    \label{equation:lagrangian}
\end{align}
where $\lambda > 0$ is the regularization coefficient. Augmented Lagrangian approaches like Eq. \eqref{equation:lagrangian} can achieve good supervised performance while minimizing the surrogate loss. 
A similar approach can be taken in the Bayesian case using posterior regularization \citep{zhu2014bayesian}, although we focus the scope of this paper on learning informative priors.

\subsection{Learning Informative Priors}
\label{sec: learning informative priors}

We present our method for incorporating domain knowledge in the form of these losses into an informative prior for BNNs. As we want to encourage sampling models that achieve low values of the surrogate loss $\phi$, our goal is to learn a prior that assigns high probability mass to these models, consequently influencing samples from the posterior.

We propose to learn our informative prior by inferring the posterior distribution over $w$ given unlabeled data $X'$ and a surrogate loss $\phi$. By Bayes' rule, this posterior is given by
\begin{align*}
    p(w | X', \phi) \propto p(\phi | w, X') \cdot p(w),
\end{align*}
where the weight-space prior $p(w) = \prod_{i} \cN(w_i; 0, \sigma_i^2)$ is the commonly used isotropic Gaussian distribution.
Since our goal is to enforce $\phi(h_w, x)$ to be small, we assume that the likelihood for $\phi$ is given by
$$
p(\phi | w, x)  = \cN\left(\phi(h_w, x); 0, \tau^2\right),
$$
where $\tau > 0$ is a hyperparameter controlling how much probability mass we want to center about models that most satisfy our domain knowledge. The posterior distribution, which represents our domain knowledge-informed prior that can be used in later tasks, is then given by
\begin{equation}\label{eq:pretrain-posterior}
p(w | X', \phi) \propto \prod_{x_i' \in X'}\cN(\phi(h_w, x_i); 0, \tau^2) \cdot p(w).
\end{equation}
Given the intractability of computing the true posterior in Eq. \eqref{eq:pretrain-posterior}, we use variational inference \citep{kingma2013auto,blei-vi} to approximate it with the low-rank multivariate Gaussian distribution
\begin{align}\label{eq:pretrain-approx-posterior}
q_\psi(w) = \cN(w; \mu, \Sigma_{r}),\quad \Sigma_r = \sum_{i=1}^r v_i v_i^T + \sigma^2 I,
\end{align}
where $\psi = (\mu, v_1,\dots, v_r)$ and where $\sigma > 0$ is a small, fixed value that keeps $\Sigma_{r}$ positive definite and $\mu$ is a vector of real-valued means. We assume that the variational covariance matrix $\Sigma_r$ has low rank $r$ for computational efficiency, given that $w$ is generally high-dimensional, which is a standard assumption in practice. As such, the size of our BNN scales as $O(r \cdot n)$, where $n$ represents the number of parameters in the neural network architecture.

\paragraph{Our Variational Objective.} We optimize the variational parameters $\psi$
to maximize the evidence lower bound (ELBO) which is given by
\begin{equation}\label{eq:variational-objective}
\mathbb{E}_{w \sim q_\psi}[\log p(\phi| w, X')] - \operatorname{KL}(q_\psi(w) || p(w)).
\end{equation}
This is a lower bound of $\log p(\phi|X')$, and optimality is achieved when $q(w) = p(w | \phi, X')$.
Since $q(w)$ and $p(w)$ are both multivariate Gaussian distributions, sampling from these distributions is straightforward, and the KL divergence term between $p(w)$ and $q(w)$ admits a closed form that can be computed efficiently. For a given set of unlabeled examples $X'$, we thus seek to optimize the objective
\begin{align*}
    \max_{\psi}  \left( \mathbb{E}_{w \sim q_\psi}\left[-\sum_{i=1}^k \frac{\phi(h_w, x)^2} {2\tau^2}\right]  -\text{KL}(q_\psi(w) || p(w))  \right).
\end{align*}
We note that as $\tau \to \infty$, we recover $q_{\psi}(w) = p(w)$. We then use the learned intermediate posterior distribution $q_\psi(w)$ as our informative prior for downstream tasks, performing posterior sampling via standard methods (e.g., SGLD, MultiSWAG) in practice.  Since our informative prior $q_{\psi}(w)$ is a low-rank Gaussian distribution, we remark that the computational overhead of approximate inference with the informative prior is similar to using an isotropic Gaussian.

\subsection{Transferring Informative Priors}

A key limitation of the learned priors is that they are architecture-specific. To make them usable for downstream tasks where other model architectures may be more suitable, it is important to identify effective techniques for \textit{transferring} these learned priors.
Our proposed strategy is to match functions drawn from the learned informative prior and a target prior distribution for the new model architecture. 

Formally, let $\cH_1 = \{h_w \mid h_w : \cX \to \cY \}$ represent the hypothesis class of our original model architecture, with a corresponding informative prior $q_{\psi_1}(w)$. We want to learn an informative prior for a different class of networks $\cH_2 = \{h_u \mid h_u : \cX \to \cY\}$. We hope to learn a distribution $q_{\psi_2}(u)$ such that the distributions over $\cH_1$ and $\cH_2$ induced by $w \sim q_{\psi_1}(w)$ and $u \sim q_{\psi_2}(u)$ are close. As we consider low-rank Gaussian priors, we can efficiently draw samples from  $q_{\psi_1}(w)$ and $q_{\psi_2}(u)$, and this motivates us to learn $\psi_2$ such that the set of functions $\{h_{w_1}, \dots, h_{w_n}\} \subseteq \cH_1$ and $\{h_{u_1}, \dots, h_{u_n}\} \subseteq \cH_2$ are \textit{similar} when $w_i \sim q_{\psi_1}(w)$, $u_i \sim q_{\psi_2}(u)$.
Since the members of each set are functions, it is difficult to compare them directly. If we have access to a set of unlabeled examples $X' = \{x_1,\dots, x_m\}$, we can instead make sure that the evaluation of each function on $X'$ are similar, i.e., $W:= \{h_{w_1}(X'), \dots, h_{w_n}(X')\}$ and $U:= \{h_{u_1}(X'), \dots, h_{u_n}(X')\}$ are similar when $h(X') = (h(x_1),\dots, h(x_m)) \in \mathbb{R}^m$. 

\paragraph{Moment Matching.} We consider simple approaches to match the moments of the two distributions $q_{\psi_1}(w)$ and $q_{\psi_2}(u)$, whose objectives are given by
\begin{align*}
    \hat{M}_1& = \mathbb{E}_{x}[(\mathbb{E}_{w \sim q_{\psi_1}(w)}[h_w(x)] -  \mathbb{E}_{u \sim q_{\psi_2}(w)}[h_u(x)])^2 ]\\
    \hat{M}_2& = \mathbb{E}_{x}[(\mathbb{E}_{w \sim q_{\psi_1}(w)}[h_w(x)^2] -  \mathbb{E}_{u \sim q_{\psi_2}(w)}[h_u(x)^2])^2 ],
\end{align*}
where $\hat{M}_1$ is used to match only the first moment, and $\hat{M}_2$ is used to match the first two moments. 

\paragraph{Maximum Mean Discrepancy.} We propose to minimize the kernel maximum mean discrepancy (MMD) \citep{gretton2012kernel, li2015generative} between $W$ and $U$, where the objective is given by
\begin{align*}
    \hat{M}(W,U) &= \frac{1}{n(n-1)}\sum_{i=1}^n\sum_{j\neq i} k( h_{w_i}(X'), h_{w_j}(X'))\\
    &+ \frac{1}{n(n-1)}\sum_{i=1}^n\sum_{j\neq i} k( h_{u_i}(X'), h_{u_j}(X'))\\
    &+\frac{1}{n^2}\sum_{i=1}^n\sum_{j=1}^n k( h_{w_i}(X'), h_{u_j}(X')),
\end{align*}

and $k$ represents a kernel. MMD only requires access to the samples from each distribution which fit well with our scenario as these samples are easy to draw. 

Meanwhile, we remark that other approaches, such as learning $\psi_2$ to fool a discriminator network that is trained to distinguish between two set of samples \citep{goodfellow2014generative, radford2015unsupervised, arjovsky2017wasserstein, li2017mmd, binkowski2018demystifying} or directly working with kernel two-sample tests for functional data \citep{wynne2022kernel}, can also be used. 
A main benefit of studying these prior transferring approaches is that they enable transferring domain knowledge when we no longer have access to the function $\phi$. This approach can help support the open-source release and usage of informative priors, similar to how pretrained models are currently used in practice.

\section{EXPERIMENTS}

We compare our method of learning an informative prior through variational inference, which we refer to as \textbf{Banana}, against 
BNN implementations with various priors, including the standard isotropic Gaussian, a Gaussian with hyperparameters optimized via empirical Bayes using Laplace's method \citep{daxberger2021laplace}, and a prior that is learned to match a GP prior with a RBF kernel \citep{tran2022all}. 
We note that the baseline matched to a GP prior (single-output) is not evaluated on our regression dataset (Pendulum), which has multivariate outputs.

For all prediction tasks described below, we use a two-layer feedforward neural network with ReLU activations. 
As noted in \citet{shwartz2022pre}, the scaling of the prior in the downstream task has a significant impact on both performance as well as the weighting of our domain knowledge. As such, we add a hyperparameter $\beta > 0$ that scales the KL divergence term in Eq. \eqref{eq:variational-objective}, to control better the tradeoff between using prior information and fitting the observed labeled data.
In computing our posterior averages for each method, we average in the logit space of the posterior samples. We also explore averaging in the output space of posterior samples in Appendix \ref{appx:model_avg}
We provide additional experimental details in Appendix \ref{sec:exp-details}.

\begin{table*}[t]
    \centering
    \caption{Comparison of Banana (with posterior averaging over logits) against BNNs with different priors in terms of accuracy, AUROC, or $L_1$ loss and $\phi$ ($\pm$ s.e.), when averaged over 5 seeds. 
    $\uparrow$ denotes that higher is better, and $\downarrow$ denotes that lower is better. We bold the method with the best performance and the lowest value of $\phi$. - denotes when a method is not applicable.}
    \setlength{\tabcolsep}{4pt}
    {\renewcommand{\arraystretch}{1.15}
    \resizebox{2.08\columnwidth}{!}{%
    \begin{tabular}{l cc cc cc} \toprule
         & \multicolumn{2}{c}{\textbf{DecoyMNIST}} &  \multicolumn{2}{c}{\textbf{MIMIC-IV}} & \multicolumn{2}{c}{\textbf{Pendulum}} \\
         \cmidrule(lr){2-3} \cmidrule(lr){4-5} \cmidrule(lr){6-7}  
         Method & Accuracy ($\uparrow$) & $\phi_{\text{background}}$ & AUROC ($\uparrow$) & $\phi_{\text{clinical}}$ & $L_1$ Loss ($\downarrow$) & $\phi_{\text{energy\_damping}}$ \\ \midrule
         BNN + Isotropic                           &  76.41 $\pm$ 0.71 & 1.06 $\pm$ 0.06  & 0.6981 $\pm$ 0.0003 & 0.1624 $\pm$ 0.0005  & \textbf{0.0036 $\pm$ 0.0001} & 0.0319 $\pm$ 0.0026 \\
         BNN + Laplace & 76.47 $\pm$ 0.70 & 1.14 $\pm$ 0.04 & 0.6980 $\pm$ 0.0002 & 0.1625 $\pm$ 0.0009 & 0.0043 $\pm$ 0.0006 & 0.0367 $\pm$ 0.0100 \\
         BNN + GP Prior & 75.49 $\pm$ 0.70 & 1.54 $\pm$ 0.04 & 0.6979 $\pm$ 0.0002 & 0.1628 $\pm$ 0.0008 & - & -\\
         \textbf{Banana}                & \textbf{78.21 $\pm$ 0.40} & \textbf{0.44 $\pm$ 0.01}  & \textbf{0.6983 $\pm$ 0.0001} & \textbf{0.1619 $\pm$ 0.0005} & 0.0041 $\pm$ 0.0007 & \textbf{0.0025 $\pm$ 0.0010}  \\ \bottomrule
    \end{tabular}
    }
    }
    \label{tab:accs_phi}
    \vspace{-0mm}
\end{table*}

\subsection{Datasets and Domain Knowledge Losses}\label{sec:data}

\paragraph{Fairness in Hiring Decisions.} We demonstrate that our method can incorporate notions of fairness on the \textbf{Folktables} dataset \citep{ding2021retiring}. We consider the task of determining whether a particular applicant gets employed, within the Alabama subset of the data in 2018. We focus on group fairness as our underlying domain knowledge, where we define our $\phi$ as
\begin{align*}
    \phi_{\text{group\_fairness}}(h, x) = \left(p(h(x) | A = a) - p(h(x) | A = b) \right)^2, 
\end{align*}
where $A$ denotes a random variable for a particular group, such as race or gender. In our experiments, we consider $A = a$ to be the subgroup that corresponds to Black people and $B = b$ to correspond to White people.
We note that satisfying this domain knowledge does not necessarily improve predictive performance \citep{pmlr-v119-dutta20a}, although it is a desirable and potentially legal necessity of a model.
    
\paragraph{Feature Importance for Image Classification.} We also demonstrate that our method can incorporate notions of feature importance. We consider the task of ignoring background information which are spurious features on a variant of the MNIST dataset \citep{lecun1998mnist} called \textbf{DecoyMNIST} \citep{ross2017right}. On this task, a patch has been added in the background that correlates with different labels at train and test time. Thus, standard supervised learning methods can struggle due to this distribution shift.
Here, we consider domain knowledge of ignoring background pixels in making predictions. We compute this as 
\begin{align*}
    \phi_{\text{background}}(h, x) = || \nabla_x h(x)||_b^2,
\end{align*}
where $b$ denotes only the feature indices that correspond to the background. On DecoyMNIST, we access these feature dimensions by looking at the uncorrupted data, which is not used during training. For other tasks, we can generate these background masks via a segmentation network.

\paragraph{Clinical Rules for Healthcare Interventions.} We demonstrate how our method can be used to incorporate clinical rules into the prior using the MIMIC-IV dataset \citep{mimic-iv}. We reproduce the binary classification task in \citet{output-bnn-2020}, where the goal is to predict whether an intervention for hypotension management (e.g., vasopressors) should be given to a patient in the intensive care unit, given a set of physiological measurements. As in \citet{output-bnn-2020}, we incorporate the clinical knowledge that an intervention should be made if the patient exhibits: (i) high lactate and low bicarbonate levels, or (ii) high creatinine levels, high blood urea nitrogen (BUN) levels, and low urine output.
We can express this knowledge as
\begin{align*}
    \phi_{\text{clinical}}(h, x) &= \mathbf{1}[x \in \mathcal{X}_c] \cdot \text{ReLU}(1-h(x)),
\end{align*}
where $h(x)$ is the classifier output, $\mathcal{X}_c \subseteq \mathcal{X}$ denotes the subset of the input space that satisfies the conditions specified in the above rules,
and $\text{ReLU}(1-h(x))$ encourages $h(x)$ on such inputs to be close to 1. 
We include all details on cohort selection, data preprocessing, and the specification of $\mathcal{X}_c$ in Appendix \ref{appendix-healthcare}.

\begin{figure}[t]
    \centering
    \includegraphics[height=2.4in]{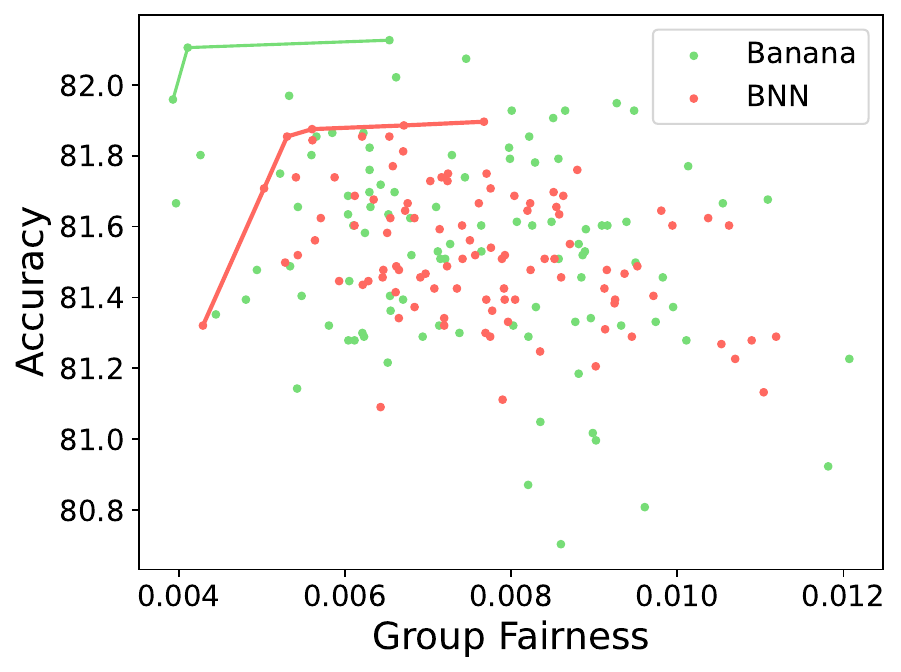}
    \caption{Comparison of samples from the Banana prior (in green) against those from an isotropic Gaussian prior (in red) on the Folktables dataset. The solid line is the Pareto frontier, and the dots represent samples from each prior. Samples from Banana tend to be more Pareto-optimal, achieving higher accuracy while satisfying fairness constraints.
    }
    \vspace{-4mm}
    \label{fig:pareto}
\end{figure}

\begin{table*}[t]
    \centering
    \caption{Our proposed methods for transferring priors successfully improve alignment with domain knowledge in BNNs with different architectures. (Top 2 rows) Values of $\phi$ for models drawn from an isotropic Gaussian prior and the learned Banana prior. (Bottom 3 rows) Values of $\phi$ from models with larger architectures drawn from an isotropic Gaussian prior and a prior transferred from Banana via a first moment matching (with SWAG) or MMD. Results are averaged over 5 seeds.}
    \label{tab:sample_prior}
    \setlength{\tabcolsep}{4pt}
    {\renewcommand{\arraystretch}{1.15}
    \resizebox{1.9\columnwidth}{!}{%
    \begin{tabular}{l cccc} \toprule
         \textbf{Method} & \textbf{DecoyMNIST} & \textbf{Folktables} & \textbf{MIMIC-IV} & \textbf{Pendulum}   \\ \midrule
Isotropic &$0.2617\pm0.0168$ & $0.0173\pm0.0056$ &$0.2867\pm0.0109$ &$134.0293\pm9.7799$ \\
\textbf{Banana}  & \textbf{0.0186 $\pm$ 0.0023} & \textbf{0.0119 $\pm$ 0.0099} & \textbf{0.0006 $\pm$ 0.001} & \textbf{0.0 $\pm$ 0.0} \\ \midrule
Isotropic (L) &$0.4963\pm0.0151$ &$0.0231\pm0.0025$ &$0.2932\pm0.0091$ &$179.6625\pm15.9685$\\
\textbf{Banana + MMD} & \textbf{0.0288 $\pm$ 0.0044} & \textbf{0.02 $\pm$ 0.0048} & \textbf{0.004 $\pm$ 0.002} & \textbf{0.0035 $\pm$ 0.0055} \\
\textbf{Banana + 1st Moment (SWAG)} & \textbf{0.0015 $\pm$ 0.0005} & \textbf{0.0014 $\pm$ 0.0021} & \textbf{0.0 $\pm$ 0.0} & \textbf{0.0 $\pm$ 0.0} \\ \bottomrule
    \end{tabular}
    }
    \vspace{-0mm}
    }
\end{table*}

\paragraph{Physics Rules for Pendulums.}
We demonstrate how our method can incorporate physical knowledge into the prior on the double pendulum dataset \citep{seo2021controlling, asseman2018learning}. We consider a regression task where the goal is to predict the next state of double-pendulum dynamics with friction from a given initial state $x= \left(\theta_{1}, \omega_{1}, \theta_2, \omega_2\right)$ where $\theta_i, \omega_i$ are the angular displacement and the velocity of the $i$-th pendulum, respectively. We incorporate physics knowledge from the law of conservation of energy; since the system has friction, the total energy of the system must be strictly decreasing over time. We can express this knowledge by the following loss:
$$\phi_{\text{energy\_damping}}(h, x) =\max(E(h(x))-E(x), 0),$$
where $h(x)$ is the predicted next state and $E(x)$ is a function that maps a given state $x$ to its total energy. This loss penalizes predictions of states with higher total energy.
\subsection{Results}\label{sec:results}
We present the results comparing our method to the baselines in Table \ref{tab:accs_phi}. We observe that incorporating the domain knowledge leads to a better-performing classifier than standard BNN approaches with existing techniques to specify priors. 
We note that across all tasks, the model averages produced by Banana achieve lower values of $\phi$ than a standard BNN approach that uses an isotropic Gaussian prior. 
We note that in some cases, such as on MIMIC-IV, the benefits in terms of $\phi$ are less clear because the domain knowledge loss $\phi_{\text{clinical}}$ is also related to the standard training objective. However, in cases where these domain knowledge losses are not as strongly connected to the training objective (i.e., the notion of patches in DecoyMNIST), the decrease in the value of $\phi$ is much clearer.
We remark that on the pendulum task, there is already sufficient labeled data (18000 samples) so the informative prior information is not necessary for strong performance. As a result, Banana is slightly less performant but better adheres to physical rules. 

On the Folktables dataset, $\phi_{\text{group\_fairness}}$ may be at odds with the underlying accuracy, i.e., a less performant model may achieve a lower value of $\phi$. As such, we omit this dataset from Table \ref{tab:accs_phi} and, rather, present the tradeoff between accuracy and group fairness. In Figure \ref{fig:pareto}, we provide a Pareto frontier of the accuracy and group fairness achieved by posterior samples from our approach, compared to those from a standard BNN. We observe that samples from our informative prior produce models that achieve lower values of group fairness, with no degradations in performance when compared to samples from an isotropic Gaussian prior.

\subsubsection{Directly Sampling from the Informative Prior}

To analyze how well the informative prior encodes our domain knowledge, we can directly sample from our informative prior and compute the value of domain knowledge loss $\phi$ achieved on our sample (see the first two rows of Table \ref{tab:sample_prior}), although we note that these models are not necessarily suited for a downstream task. We use the same hyperparameter values for training our informative prior as those selected for the downstream classification/regression task in Table \ref{tab:accs_phi}. On each dataset, we compute our expected value of $\phi$ over 10 samples from the informative prior.
We observe that across almost every task, our informative prior successfully upweights models that achieve significantly lower values of $\phi$ on their respective datasets when compared to randomly sampling from an isotropic Gaussian distribution, reflected by a posterior average that has much smaller values of $\phi$.

\begin{figure}[t]
    \centering
    \includegraphics[height=2.3in]{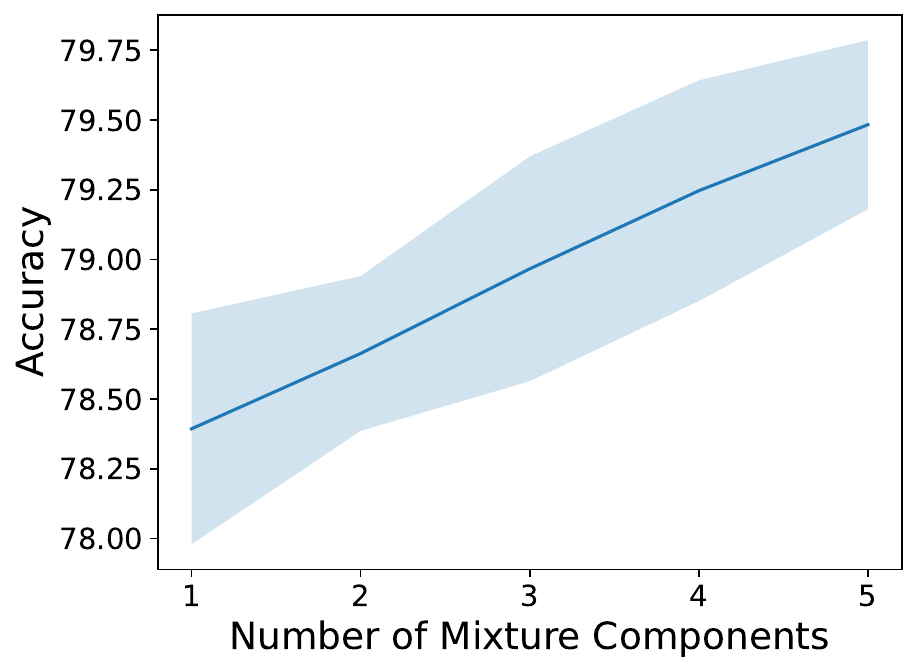}
    \caption{Change in test accuracy on the DecoyMNIST task when varying the number of mixture components in the informative prior in Banana. Results are averaged over 5 seeds, and the shaded region represents mean $\pm$ s.e.}
    \label{fig:multiswag}
    \vspace{-2mm}
\end{figure}

\subsubsection{Transferring Priors to Different Architectures}

In addition to comparing the value of $\phi$ achieved by models sampled from our informative prior, we also evaluate the performance of transferring this informative prior to a different model architecture (see the bottom three rows in Table \ref{tab:sample_prior}). We transfer the prior over a two-layer neural network to another two-layer network with a larger hidden dimension size, where we minimize the difference between first moments (via SWAG \citep{maddox2019simple}) or MMD with respect to a Gaussian kernel. 

We observe that our transferring approach yields informative priors over the new model class that also reflect much smaller values of $\phi$ than standard isotropic Gaussian prior over the larger model architecture, almost fully recovering the same performance as the original prior in many cases. We further compare against other strategies that we propose to transfer this prior in additional ablations in Appendix \ref{appendix:prior_transfer}. These results demonstrate that informative priors learned in Banana can effectively be transferred to different model architectures that may be better suited for the downstream task.

\subsection{Ablation Studies}\label{sec:ablation}

\paragraph{Alternative Approximations for the Informative Prior.}
We assess different approximation schemes for learning the informative prior to better capture the knowledge in $\phi$. Given that variational inference often underestimates the variance of the true posterior \citep{blei-vi}, which need not be unimodal, we consider approximating Eq. \eqref{eq:pretrain-posterior} with a \textit{mixture} $q(w) = \frac{1}{K} \sum_{k=1}^K \mathcal{N}(\mu_k,\Sigma_{r,k})$ of $K$ rank-$r$ Gaussians, via the MultiSWAG method \citep{bdl}.
For each $k=1,\ldots,K$, we initialize the model parameters with a different random seed and compute $(\mu_k,\Sigma_{r,k})$ by averaging over 3 samples (5 epochs apart) from the stochastic gradient descent (SGD) trajectory, after an initial 5 epochs of training. Using each $q_k$ as an informative prior, we sample 5 weights from the downstream posterior via SGLD. We average over sampled weights to obtain the final prediction. 

Figure \ref{fig:multiswag} shows the change in test accuracy on DecoyMNIST as we vary $K$ from 1 to 5. We find that on average, the test accuracy monotonically increases with increasing $K$ and improves over the test accuracy in Table \ref{tab:accs_phi}. These results suggest that with a sufficient computational budget, learning a multimodal informative prior can be an effective approach for better capturing our domain knowledge and improving downstream performance. This also shows that a sufficiently complex $q$ is required to fully reap the benefits of using prior information. We study this further in Appendix \ref{appx:prior_complexity}, where we experiment with higher rank approximations of our prior as well as a diagonalized Gaussian prior approximation.

\paragraph{Model Ensemble Size.}

With other frequentist methods, computing a model ensemble approximately equivalent to a Bayesian model average \citep{deep-ensembles} requires performing potentially computationally expensive regularization with $\phi$ (or even pretraining in the case of invariances used in self-supervised learning as the form of domain knowledge). On the other hand, a Bayesian approach only needs to train the prior \textit{once} and can generate multiple samples from the posterior. Given that the computational cost of sampling from the posterior is efficient as in our setting, this can be a significant benefit, specifically when computing this regularization or pretraining is costly.

\begin{figure}[t]
    \centering
    \includegraphics[height=2.3in]{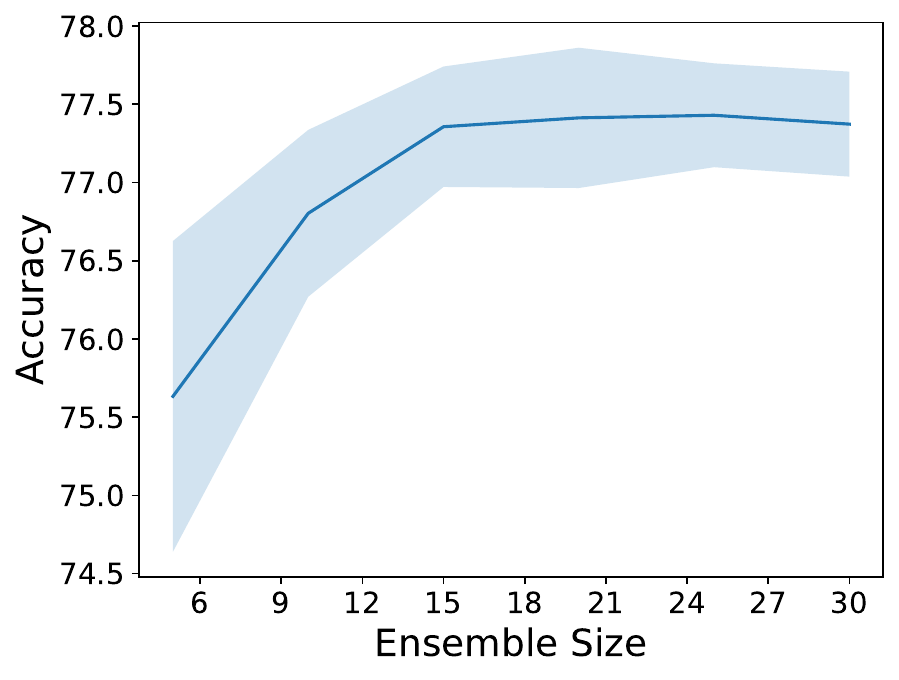}
    \caption{Change in test accuracy on the DecoyMNIST task when varying the number of models sampled to compute our posterior average in Banana. Results are averaged over 5 seeds, and the shaded region represents mean $\pm$ s.e.}
    \label{fig:ensemble_ablation}
    \vspace{-2mm}
\end{figure}

As such, we run ablation studies to evaluate the performance of Banana's and other approaches' ensembles to assess how the number of members of the ensemble impacts the downstream performance of the approaches. We observe that increasing the ensemble size increases performance, until we observe diminishing returns after ensemble sizes of 15, on the DecoyMNIST task (Figure \ref{fig:ensemble_ablation}). This demonstrates that larger ensembles generally achieve better performance and supports the use of informative priors, which can more efficiently scale to posterior averages with larger ensembles when compared to other regularization-based approaches. 

\section{DISCUSSION}

We propose a framework to incorporate general forms of domain knowledge into the priors for BNNs. Empirically, we observe that this can improve the performance of BNNs across several tasks with different notions of domain knowledge and leads to models that exhibit desirable properties. In addition, we provide an effective approach to transfer informative priors across model architectures, resolving an existing problem in the literature. Our results provide new insights into incorporating domain knowledge into priors for Bayesian methods, which can be captured by optimizing a learnable approximation through variational inference. 

As a whole, our results support the development of open-source informative priors that practitioners can incorporate into their various specific use cases to encode desirable model properties without the need to deal with $\phi$ directly, irrespective of the desired model architecture. As such, this supports  
the foundations for pretraining in the Bayesian setting, by providing a framework to develop and transfer informative priors to new model tasks as desired, similar to pretrained weights or foundation models that are currently released as open-source.

\section*{Acknowledgements}

DS was supported by the Bosch Center for Artificial Intelligence, the ARCS Foundation. This material is based upon work supported by the National Science Foundation Graduate Research Fellowship under Grant No. DGE2140739.
RP was supported by a Bloomberg Data Science PhD fellowship.
DPJ was supported by DARPA via FA8750-23-2-1015 and ONR via N00014-23-1-2368. 
YB was supported in part by the AI2050 program at Schmidt Sciences (Grant G-22-64474). 
DPJ and YB also gratefully acknowledge the NSF (IIS2211955), UPMC, Highmark Health, Abridge, Ford Research, Mozilla, the PwC Center, Amazon AI, JP Morgan Chase, the Block Center, the Center for Machine Learning and Health, and the CMU Software Engineering Institute (SEI) via Department of Defense contract FA8702-15-D-0002, for their generous support of ACMI Lab's research.
\bibliography{ref}

\begin{thebibliography}{59}
\providecommand{\natexlab}[1]{#1}
\providecommand{\url}[1]{\texttt{#1}}
\expandafter\ifx\csname urlstyle\endcsname\relax
  \providecommand{\doi}[1]{doi: #1}\else
  \providecommand{\doi}{doi: \begingroup \urlstyle{rm}\Url}\fi

\bibitem[Arjovsky et~al.(2017)Arjovsky, Chintala, and Bottou]{arjovsky2017wasserstein}
Martin Arjovsky, Soumith Chintala, and L{\'e}on Bottou.
\newblock {Wasserstein Generative Adversarial Networks}.
\newblock In \emph{International Conference on Machine Learning}, pages 214--223. PMLR, 2017.

\bibitem[Asseman et~al.(2018)Asseman, Kornuta, and Ozcan]{asseman2018learning}
Alexis Asseman, Tomasz Kornuta, and Ahmet Ozcan.
\newblock {Learning Beyond Simulated Physics}.
\newblock \emph{In Modeling and Decision-Making in the Spatiotemporal Domain Workshop}, 2018.

\bibitem[Bi{\'n}kowski et~al.(2018)Bi{\'n}kowski, Sutherland, Arbel, and Gretton]{binkowski2018demystifying}
Miko{\l}aj Bi{\'n}kowski, Danica~J Sutherland, Michael Arbel, and Arthur Gretton.
\newblock {Demystifying MMD GANs}.
\newblock In \emph{International Conference on Learning Representations}, 2018.

\bibitem[Blei et~al.(2017)Blei, Kucukelbir, and McAuliffe]{blei-vi}
David~M. Blei, Alp Kucukelbir, and Jon~D. McAuliffe.
\newblock {Variational Inference: A Review for Statisticians}.
\newblock \emph{Journal of the American Statistical Association}, 112\penalty0 (518):\penalty0 859--877, 2017.

\bibitem[Chen et~al.(2014)Chen, Fox, and Guestrin]{chen2014stochastic}
Tianqi Chen, Emily Fox, and Carlos Guestrin.
\newblock {Stochastic Gradient Hamiltonian Monte Carlo}.
\newblock In \emph{International Conference on Machine Learning}, 2014.

\bibitem[Chen et~al.(2020)Chen, Kornblith, Norouzi, and Hinton]{simclr}
Ting Chen, Simon Kornblith, Mohammad Norouzi, and Geoffrey Hinton.
\newblock {A Simple Framework for Contrastive Learning of Visual Representations}.
\newblock In \emph{International Conference on Machine Learning}, 2020.

\bibitem[Daxberger et~al.(2021)Daxberger, Kristiadi, Immer, Eschenhagen, Bauer, and Hennig]{daxberger2021laplace}
Erik Daxberger, Agustinus Kristiadi, Alexander Immer, Runa Eschenhagen, Matthias Bauer, and Philipp Hennig.
\newblock {Laplace Redux: Effortless Bayesian Deep Learning}.
\newblock \emph{Advances in Neural Information Processing Systems}, 34:\penalty0 20089--20103, 2021.

\bibitem[de~Avila Belbute-Peres et~al.(2018)de~Avila Belbute-Peres, Smith, Allen, Tenenbaum, and Kolter]{NEURIPS2018_842424a1}
Filipe de~Avila Belbute-Peres, Kevin Smith, Kelsey Allen, Josh Tenenbaum, and J.~Zico Kolter.
\newblock {End-to-End Differentiable Physics for Learning and Control}.
\newblock In \emph{Advances in Neural Information Processing Systems}, 2018.

\bibitem[Ding et~al.(2021)Ding, Hardt, Miller, and Schmidt]{ding2021retiring}
Frances Ding, Moritz Hardt, John Miller, and Ludwig Schmidt.
\newblock {Retiring Adult: New Datasets for Fair Machine Learning}.
\newblock \emph{Advances in Neural Information Processing Systems}, 2021.

\bibitem[Dutta et~al.(2020)Dutta, Wei, Yueksel, Chen, Liu, and Varshney]{pmlr-v119-dutta20a}
Sanghamitra Dutta, Dennis Wei, Hazar Yueksel, Pin-Yu Chen, Sijia Liu, and Kush Varshney.
\newblock {Is There a Trade-Off Between Fairness and Accuracy? {A} Perspective Using Mismatched Hypothesis Testing}.
\newblock In \emph{International Conference on Machine Learning}, 2020.

\bibitem[Dwork et~al.(2012)Dwork, Hardt, Pitassi, Reingold, and Zemel]{dwork2012fairness}
Cynthia Dwork, Moritz Hardt, Toniann Pitassi, Omer Reingold, and Richard Zemel.
\newblock {Fairness Through Awareness}.
\newblock In \emph{Innovations in Theoretical Computer Science Conference}, 2012.

\bibitem[Fortuin(2022)]{fortuin-priors}
Vincent Fortuin.
\newblock {Priors in Bayesian Deep Learning: A Review}.
\newblock \emph{International Statistical Review}, 90\penalty0 (3):\penalty0 563--591, 2022.

\bibitem[Gelman et~al.(2017)Gelman, Simpson, and Betancourt]{gelman-2017}
Andrew Gelman, Daniel Simpson, and Michael Betancourt.
\newblock {The Prior Can Often Only Be Understood in the Context of the Likelihood}.
\newblock \emph{Entropy}, 19\penalty0 (10), 2017.

\bibitem[Gelman et~al.(2020)Gelman, Vehtari, Simpson, Margossian, Carpenter, Yao, Kennedy, Gabry, Bürkner, and Modrák]{gelman2020bayesian}
Andrew Gelman, Aki Vehtari, Daniel Simpson, Charles~C. Margossian, Bob Carpenter, Yuling Yao, Lauren Kennedy, Jonah Gabry, Paul-Christian Bürkner, and Martin Modrák.
\newblock {Bayesian Workflow}.
\newblock \emph{arXiv preprint arXiv:2011.01808}, 2020.

\bibitem[Goodfellow et~al.(2014)Goodfellow, Pouget-Abadie, Mirza, Xu, Warde-Farley, Ozair, Courville, and Bengio]{goodfellow2014generative}
Ian Goodfellow, Jean Pouget-Abadie, Mehdi Mirza, Bing Xu, David Warde-Farley, Sherjil Ozair, Aaron Courville, and Yoshua Bengio.
\newblock {Generative Adversarial Nets}.
\newblock volume~27, 2014.

\bibitem[Gretton et~al.(2012)Gretton, Borgwardt, Rasch, Sch{\"o}lkopf, and Smola]{gretton2012kernel}
Arthur Gretton, Karsten~M Borgwardt, Malte~J Rasch, Bernhard Sch{\"o}lkopf, and Alexander Smola.
\newblock {A Kernel Two-Sample Test}.
\newblock \emph{The Journal of Machine Learning Research}, 13\penalty0 (1):\penalty0 723--773, 2012.

\bibitem[Gr{\"u}nwald and van Ommen(2017)]{prior-misspecification}
Peter Gr{\"u}nwald and Thijs van Ommen.
\newblock {Inconsistency of Bayesian Inference for Misspecified Linear Models, and a Proposal for Repairing It}.
\newblock \emph{Bayesian Analysis}, 12\penalty0 (4):\penalty0 1069 -- 1103, 2017.

\bibitem[Hafner et~al.(2019)Hafner, Tran, Lillicrap, Irpan, and Davidson]{noise-contrastive-priors}
Danijar Hafner, Dustin Tran, Timothy~P. Lillicrap, Alex Irpan, and James Davidson.
\newblock {Noise Contrastive Priors for Functional Uncertainty}.
\newblock In \emph{Uncertainty in Artificial Intelligence}, 2019.

\bibitem[Hardt et~al.(2016)Hardt, Price, and Srebro]{hardt2016equality}
Moritz Hardt, Eric Price, and Nati Srebro.
\newblock {Equality of Opportunity in Supervised Learning}.
\newblock \emph{Advances in Neural Information Processing Systems}, 2016.

\bibitem[Henaff(2020)]{henaff-2020}
Olivier Henaff.
\newblock {Data-Efficient Image Recognition with Contrastive Predictive Coding}.
\newblock In \emph{International Conference on Machine Learning}, 2020.

\bibitem[Hendrycks and Gimpel(2016)]{hendrycks2016baseline}
Dan Hendrycks and Kevin Gimpel.
\newblock {A Baseline for Detecting Misclassified and Out-of-Distribution Examples in Neural Networks}.
\newblock In \emph{International Conference on Learning Representations}, 2016.

\bibitem[Hendrycks et~al.(2021)Hendrycks, Basart, Mu, Kadavath, Wang, Dorundo, Desai, Zhu, Parajuli, Guo, et~al.]{hendrycks2021many}
Dan Hendrycks, Steven Basart, Norman Mu, Saurav Kadavath, Frank Wang, Evan Dorundo, Rahul Desai, Tyler Zhu, Samyak Parajuli, Mike Guo, et~al.
\newblock {The Many Faces of Robustness: A Critical Analysis of Out-of-Distribution Generalization}.
\newblock In \emph{International Conference on Computer Vision}, 2021.

\bibitem[Hern\'{a}ndez-Lobato and Adams(2015)]{hernandez-2015}
Jos\'{e}~Miguel Hern\'{a}ndez-Lobato and Ryan~P. Adams.
\newblock {Probabilistic Backpropagation for Scalable Learning of Bayesian Neural Networks}.
\newblock In \emph{International Conference on Machine Learning}, 2015.

\bibitem[Huang et~al.(2023)Huang, Pang, Liu, and Yan]{huang2023posterior}
Jiayu Huang, Yutian Pang, Yongming Liu, and Hao Yan.
\newblock {Posterior Regularized Bayesian Neural Network Incorporating Soft and Hard Knowledge Constraints}.
\newblock \emph{Knowledge-Based Systems}, 259:\penalty0 110043, 2023.

\bibitem[Ismail et~al.(2021)Ismail, Corrada~Bravo, and Feizi]{ismail2021improving}
Aya~Abdelsalam Ismail, Hector Corrada~Bravo, and Soheil Feizi.
\newblock {Improving Deep Learning Interpretability by Saliency Guided Training}.
\newblock \emph{Advances in Neural Information Processing Systems}, 2021.

\bibitem[Johnson et~al.(2023)Johnson, Bulgarelli, Shen, Gayles, Shammout, Horng, Pollard, Hao, Moody, Gow, Lehman, Celi, and Mark]{mimic-iv}
Alistair E.~W. Johnson, Lucas Bulgarelli, Lu~Shen, Alvin Gayles, Ayad Shammout, Steven Horng, Tom~J. Pollard, Sicheng Hao, Benjamin Moody, Brian Gow, Li-wei~H. Lehman, Leo~A. Celi, and Roger~G. Mark.
\newblock {MIMIC-IV, A Freely Accessible Electronic Health Record Dataset}.
\newblock \emph{Scientific Data}, 10\penalty0 (1), 2023.

\bibitem[Kingma and Welling(2013)]{kingma2013auto}
Diederik~P Kingma and Max Welling.
\newblock {Auto-Encoding Variational Bayes}.
\newblock \emph{arXiv preprint arXiv:1312.6114}, 2013.

\bibitem[Lakshminarayanan et~al.(2017)Lakshminarayanan, Pritzel, and Blundell]{deep-ensembles}
Balaji Lakshminarayanan, Alexander Pritzel, and Charles Blundell.
\newblock {Simple and Scalable Predictive Uncertainty Estimation using Deep Ensembles}.
\newblock In \emph{Advances in Neural Information Processing Systems}, 2017.

\bibitem[Larson et~al.(2016)Larson, Angwin, Kirchner, and Mattu]{compas}
Jeff Larson, Julia Angwin, Lauren Kirchner, and Surya Mattu.
\newblock {How We Analyzed the COMPAS Recidivism Algorithm}.
\newblock \emph{ProPublica}, May 2016.

\bibitem[LeCun et~al.(1998)LeCun, Cortes, and Burges]{lecun1998mnist}
Yann LeCun, Corinna Cortes, and Christopher~J.C. Burges.
\newblock {The MNIST Database of Handwritten Digits}, 1998.
\newblock URL \url{{http://yann.lecun.com/exdb/mnist/}}.

\bibitem[Li et~al.(2017)Li, Chang, Cheng, Yang, and P{\'o}czos]{li2017mmd}
Chun-Liang Li, Wei-Cheng Chang, Yu~Cheng, Yiming Yang, and Barnab{\'a}s P{\'o}czos.
\newblock {MMD GAN: Towards Deeper Understanding of Moment Matching Network}.
\newblock \emph{Advances in Neural Information Processing Systems}, 30, 2017.

\bibitem[Li et~al.(2015)Li, Swersky, and Zemel]{li2015generative}
Yujia Li, Kevin Swersky, and Rich Zemel.
\newblock {Generative Moment Matching Networks}.
\newblock In \emph{International Conference on Machine Learning}, 2015.

\bibitem[MacKay(1995)]{mackay-1995}
David J~C MacKay.
\newblock {Probable Networks and Plausible Predictions — A Review of Practical Bayesian Methods for Supervised Neural Networks}.
\newblock \emph{Network: Computation in Neural Systems}, 6\penalty0 (3):\penalty0 469--505, 1995.

\bibitem[MacKay(1992)]{mackay-thesis}
David John~Cameron MacKay.
\newblock \emph{Bayesian Methods for Adaptive Models}.
\newblock PhD thesis, California Institute of Technology, 1992.

\bibitem[Maddox et~al.(2019)Maddox, Izmailov, Garipov, Vetrov, and Wilson]{maddox2019simple}
Wesley~J Maddox, Pavel Izmailov, Timur Garipov, Dmitry~P Vetrov, and Andrew~Gordon Wilson.
\newblock {A Simple Baseline for Bayesian Uncertainty in Deep Learning}.
\newblock \emph{Advances in Neural Information Processing Systems}, 2019.

\bibitem[Mehrabi et~al.(2021)Mehrabi, Morstatter, Saxena, Lerman, and Galstyan]{mehrabi2021survey}
Ninareh Mehrabi, Fred Morstatter, Nripsuta Saxena, Kristina Lerman, and Aram Galstyan.
\newblock {A Survey on Bias and Fairness in Machine Learning}.
\newblock \emph{ACM Computing Surveys (CSUR)}, 54\penalty0 (6):\penalty0 1--35, 2021.

\bibitem[Nalisnick et~al.(2021)Nalisnick, Gordon, and Miguel Hernandez-Lobato]{pred-complexity-priors}
Eric Nalisnick, Jonathan Gordon, and Jose Miguel Hernandez-Lobato.
\newblock {Predictive Complexity Priors}.
\newblock In \emph{International Conference on Artificial Intelligence and Statistics}, 2021.

\bibitem[Nalisnick(2018)]{nalisnick-2018}
Eric~T. Nalisnick.
\newblock \emph{On Priors for Bayesian Neural Networks}.
\newblock PhD thesis, University of California, Irvine, 2018.

\bibitem[Neal(1996)]{bnn-neal}
Radford~M. Neal.
\newblock \emph{Bayesian Learning for Neural Networks}.
\newblock PhD thesis, University of Toronto, 1996.

\bibitem[Obermeyer et~al.(2019)Obermeyer, Powers, Vogeli, and Mullainathan]{obermeyer}
Ziad Obermeyer, Brian Powers, Christine Vogeli, and Sendhil Mullainathan.
\newblock {Dissecting Racial Bias in An Algorithm Used to Manage the Health of Populations}.
\newblock \emph{Science}, 366\penalty0 (6464):\penalty0 447--453, 2019.

\bibitem[Pukdee et~al.(2023)Pukdee, Sam, Kolter, Balcan, and Ravikumar]{pukdee2023learning}
Rattana Pukdee, Dylan Sam, J~Zico Kolter, Nina Balcan, and Pradeep~Kumar Ravikumar.
\newblock {Learning with Explanation Constraints}.
\newblock In \emph{Advances in Neural Information Processing Systems}, 2023.

\bibitem[Radford et~al.(2016)Radford, Metz, and Chintala]{radford2015unsupervised}
Alec Radford, Luke Metz, and Soumith Chintala.
\newblock {Unsupervised Representation Learning with Deep Convolutional Generative Adversarial Networks}.
\newblock 2016.

\bibitem[Rasmussen and Williams(2005)]{gpforml}
Carl~Edward Rasmussen and Christopher K.~I. Williams.
\newblock \emph{Gaussian Processes for Machine Learning (Adaptive Computation and Machine Learning)}.
\newblock The MIT Press, 2005.

\bibitem[Rieger et~al.(2020)Rieger, Singh, Murdoch, and Yu]{rieger2020interpretations}
Laura Rieger, Chandan Singh, William Murdoch, and Bin Yu.
\newblock {Interpretations Are Useful: Penalizing Explanations to Align Neural Networks with Prior Knowledge}.
\newblock In \emph{International Conference on Machine Learning}, 2020.

\bibitem[Ross et~al.(2017)Ross, Hughes, and Doshi-Velez]{ross2017right}
Andrew~Slavin Ross, Michael~C Hughes, and Finale Doshi-Velez.
\newblock {Right for the Right Reasons: Training Differentiable Models By Constraining Their Explanations}.
\newblock In \emph{International Joint Conference on Artificial Intelligence}, 2017.

\bibitem[Sam and Kolter(2022)]{sam2022losses}
Dylan Sam and J~Zico Kolter.
\newblock {Losses over Labels: Weakly Supervised Learning via Direct Loss Construction}.
\newblock \emph{arXiv preprint arXiv:2212.06921}, 2022.

\bibitem[Seo et~al.(2021)Seo, Arik, Yoon, Zhang, Sohn, and Pfister]{seo2021controlling}
Sungyong Seo, Sercan Arik, Jinsung Yoon, Xiang Zhang, Kihyuk Sohn, and Tomas Pfister.
\newblock {Controlling Neural Networks with Rule Representations}.
\newblock \emph{Advances in Neural Information Processing Systems}, 2021.

\bibitem[Sharma et~al.(2023)Sharma, Rainforth, Teh, and Fortuin]{sharma2023incorporating}
Mrinank Sharma, Tom Rainforth, Yee~Whye Teh, and Vincent Fortuin.
\newblock {Incorporating Unlabelled Data into Bayesian Neural Networks}.
\newblock \emph{arXiv preprint arXiv:2304.01762}, 2023.

\bibitem[Shwartz-Ziv et~al.(2022)Shwartz-Ziv, Goldblum, Souri, Kapoor, Zhu, LeCun, and Wilson]{shwartz2022pre}
Ravid Shwartz-Ziv, Micah Goldblum, Hossein Souri, Sanyam Kapoor, Chen Zhu, Yann LeCun, and Andrew~G Wilson.
\newblock {Pre-Train Your Loss: Easy Bayesian Transfer Learning with Informative Priors}.
\newblock \emph{Advances in Neural Information Processing Systems}, 2022.

\bibitem[Sun et~al.(2017)Sun, Chen, and Carin]{pmlr-v54-sun17b}
Shengyang Sun, Changyou Chen, and Lawrence Carin.
\newblock {Learning Structured Weight Uncertainty in Bayesian Neural Networks}.
\newblock In \emph{International Conference on Artificial Intelligence and Statistics}, 2017.

\bibitem[Sun et~al.(2019)Sun, Zhang, Shi, and Grosse]{sun2018functional}
Shengyang Sun, Guodong Zhang, Jiaxin Shi, and Roger Grosse.
\newblock {Functional Variational Bayesian Neural Networks}.
\newblock In \emph{International Conference on Learning Representations}, 2019.

\bibitem[Tran et~al.(2022)Tran, Rossi, Milios, and Filippone]{tran2022all}
Ba-Hien Tran, Simone Rossi, Dimitrios Milios, and Maurizio Filippone.
\newblock {All You Need is a Good Functional Prior for Bayesian Deep Learning}.
\newblock \emph{The Journal of Machine Learning Research}, 23\penalty0 (1):\penalty0 3210--3265, 2022.

\bibitem[Wang et~al.(2020)Wang, McDermott, Chauhan, Ghassemi, Hughes, and Naumann]{mimic-extract}
Shirly Wang, Matthew B.~A. McDermott, Geeticka Chauhan, Marzyeh Ghassemi, Michael~C. Hughes, and Tristan Naumann.
\newblock {MIMIC-Extract: A Data Extraction, Preprocessing, and Representation Pipeline for MIMIC-III}.
\newblock In \emph{Conference on Health, Inference, and Learning}, 2020.

\bibitem[Welling and Teh(2011)]{welling2011bayesian}
Max Welling and Yee~W Teh.
\newblock {Bayesian Learning via Stochastic Gradient Langevin Dynamics}.
\newblock In \emph{International Conference on Machine Learning}, 2011.

\bibitem[Wilson and Izmailov(2020)]{bdl}
Andrew~Gordon Wilson and Pavel Izmailov.
\newblock {Bayesian Deep Learning and a Probabilistic Perspective of Generalization}.
\newblock \emph{Advances in Neural Information Processing Systems}, 2020.

\bibitem[Wynne and Duncan(2022)]{wynne2022kernel}
George Wynne and Andrew~B Duncan.
\newblock {A Kernel Two-Sample Test for Functional Data}.
\newblock \emph{The Journal of Machine Learning Research}, 23\penalty0 (1):\penalty0 3159--3209, 2022.

\bibitem[Yang et~al.(2020)Yang, Lorch, Graule, Lakkaraju, and Doshi-Velez]{output-bnn-2020}
Wanqian Yang, Lars Lorch, Moritz Graule, Himabindu Lakkaraju, and Finale Doshi-Velez.
\newblock {Incorporating Interpretable Output Constraints in Bayesian Neural Networks}.
\newblock In \emph{Advances in Neural Information Processing Systems}, 2020.

\bibitem[Zafar et~al.(2017)Zafar, Valera, Rogriguez, and Gummadi]{zafar2017fairness}
Muhammad~Bilal Zafar, Isabel Valera, Manuel~Gomez Rogriguez, and Krishna~P Gummadi.
\newblock {Fairness Constraints: Mechanisms for Fair Classification}.
\newblock In \emph{International Conference on Artificial Intelligence and Statistics}, 2017.

\bibitem[Zhu et~al.(2014)Zhu, Chen, and Xing]{zhu2014bayesian}
Jun Zhu, Ning Chen, and Eric~P Xing.
\newblock {Bayesian Inference with Posterior Regularization and Applications to Infinite Latent SVMs}.
\newblock \emph{The Journal of Machine Learning Research}, 15\penalty0 (1):\penalty0 1799--1847, 2014.

\end{thebibliography}


\onecolumn

\title{Bayesian Neural Networks with Domain Knowledge Priors\\(Supplementary Material)}
\maketitle

\renewcommand{\thesection}{\Alph{section}}
\setcounter{section}{0}

\renewcommand\thetable{A\arabic{table}}
\renewcommand\thefigure{A\arabic{figure}}
\setcounter{table}{0}
\setcounter{figure}{0}
\section{Additional Experiments}

\subsection{Model Averaging}\label{appx:model_avg}

We provide experiments to compare averaging different samples from the posterior distribution in their logit space and in their prediction space. We remark that the Pendulum dataset consists of a regression task, where there is no distinction between logit space and prediction space and, thus, we do not report those results as they are the same. We also note that it is common to average in weight space; this performs quite poorly since we do not control the norms of each layer (i.e., there are no layer norm operations).

\begin{table*}[h]
    \centering
    \caption{A comparison of different ensembling techniques of models sampled from the posterior distribution of Banana. We report accuracy ($\pm$ s.e.) when averaged over 5 seeds. }
    \normalsize
    \setlength{\tabcolsep}{4pt}
    {\renewcommand{\arraystretch}{1.1}
    \resizebox{0.6\columnwidth}{!}{%
    \begin{tabular}{l ccc} \toprule
         & \textbf{DecoyMNIST} & \textbf{Folktables} & \textbf{MIMIC-IV} \\ \midrule
         Banana - Predictions   &   77.51 $\pm$ 0.39  & 81.45 $\pm$ 0.03 & 0.6951 $\pm$ 0.0014 \\
         Banana - Logits        &   78.21 $\pm$ 0.40 & 81.50 $\pm$ 0.04 & 0.6983 $\pm$ 0.0001 \\ \bottomrule    
    \end{tabular}
    }
    }
    \label{tab:accs_phi_lag_logits}
\end{table*}

We note that there is not a significant difference, although we observe that performing model averaging over the logits of each sample from the posterior distribution achieves potentially higher accuracy than averaging over the discrete predictions (where tiebreaks are simply taken by the first class in the ordering).

\begin{figure}[t]
    \centering
    \includegraphics[width=0.4\columnwidth]{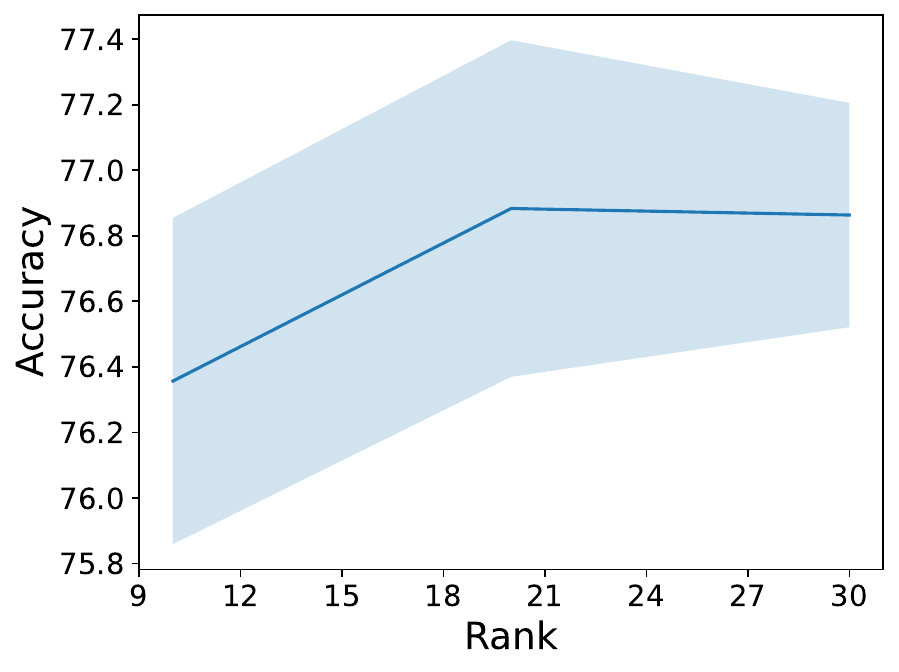}
    \caption{Results when varying the rank to approximate our informative prior in Banana on the DecoyMNIST task. Results are averaged over 5 seeds, and the shaded region represents mean $\pm$ s.e. 
    }
    \label{fig:ensemble_rank}
\end{figure}

\subsection{Varying the Complexity of Our Informative Prior Approximation} \label{appx:prior_complexity}

As demonstrated in Table \ref{tab:sample_prior}, our approach can capture domain knowledge in the form of $\phi$ through a rank-$r$ approximation of the covariance matrix of a multivariate Gaussian distribution. Here, we run ablations to study how the rank of our approximation influences downstream performance, albeit while suffering slightly larger computational costs (i.e., $O(r n)$ where $r$ is the rank and $n$ is the number of parameters). 

We observe that increasing the rank of our prior approximation on the DecoyMNIST task seems to slightly improve performance, with larger rank approximations plateuing in performance after $r = 20$. (Figure \ref{fig:ensemble_rank}). This slight increase demonstrates that learning informative priors with strong performance suffices with a small rank approximation, which is not too computationally expensive.




\begin{table*}[h]
    \centering
    \caption{Comparison of Banana (with posterior averaging over logits) and a weaker version of Banana where we only learn the means and variance of a diagonalized Gaussian approximation of the informative prior. We compare against BNNs with isotropic Gaussian priors in terms of accuracy, AUROC, or $L_1$ loss and $\phi$ ($\pm$ s.e.) when averaged over 5 seeds. $\uparrow$ denotes a metric where higher is better, and $\downarrow$ denotes that lower is better. We bold either BNN or Banana when it achieves the best performance and the lowest value of $\phi$. }
    \setlength{\tabcolsep}{2pt}
    {\renewcommand{\arraystretch}{1.1}
    \resizebox{0.95\columnwidth}{!}{%
    \begin{tabular}{l cc cc cc} \toprule
         & \multicolumn{2}{c}{\textbf{DecoyMNIST}} &  \multicolumn{2}{c}{\textbf{MIMIC-IV}} & \multicolumn{2}{c}{\textbf{Pendulum}} \\
         \cmidrule(lr){2-3} \cmidrule(lr){4-5} \cmidrule(lr){6-7}  
         Method & Accuracy ($\uparrow$) & $\phi_{\text{background}}$ & AUROC ($\uparrow$) & $\phi_{\text{clinical}}$ & $L_1$ Loss ($\downarrow$) & $\phi_{\text{energy\_damping}}$ \\ \midrule
          BNN                            &  76.41 $\pm$ 0.71 & 1.06 $\pm$ 0.06  & 0.6981 $\pm$ 0.0003 & 0.1624 $\pm$ 0.0005  & \textbf{0.0036 $\pm$ 0.0001} & 0.0319 $\pm$ 0.0026 \\
         \textbf{Banana (diag)}         & \textbf{76.52 $\pm$ 0.72} & \textbf{0.37 $\pm$ 0.01} & 0.6977 $\pm$ 0.0003 & 0.1628 $\pm$ 0.0012  & 0.0046 $\pm$ 0.0005 & \textbf{0.0170 $\pm$ 0.0050}  \\
         \textbf{Banana}                & \textbf{78.21 $\pm$ 0.40} & \textbf{0.44 $\pm$ 0.01}  & \textbf{0.6983 $\pm$ 0.0001} & \textbf{0.1619 $\pm$ 0.0005} & 0.0041 $\pm$ 0.0007 & \textbf{0.0025 $\pm$ 0.0010}  \\ \bottomrule
    \end{tabular}
    }
    }
    \label{tab:accs_phi_diag}
\end{table*}

We also provide additional experiments that use a diagonalized Gaussian approximation for this informative prior. Here, our learnable parameters are simply the means and a vector of variances along the diagonal of a diagonalized covariance matrix. We observe that this successfully captures \textit{some} information from $\phi$, but our more complex approximation via a low-rank Gaussian performs better in capturing our domain knowledge. This demonstrates that indeed a sufficiently complex prior is required to gain the full benefits of using domain knowledge.

\subsection{Comparison against Other Prior Transfer Techniques} \label{appendix:prior_transfer}

We provide comparisons against additional techniques to transfer the prior learned in Banana across different model architectures. We observe that MMD and 1st Moment Matching using SWAG \citep{maddox2019simple} perform favorably when compared to simply matching and directly optimizing over the learnable parameters of the prior approximation.

\begin{table*}[h]
    \centering
    \caption{Comparing the value of $\phi$ of models drawn from an isotropic Gaussian prior and an informative prior transferred from Banana to a larger network size in terms of hidden dimension size via multiple moment matching techniques and MMD. Results are averaged over 5 seeds.}
    \setlength{\tabcolsep}{4pt}
    {\renewcommand{\arraystretch}{1.1}
    \resizebox{0.95\columnwidth}{!}{%
    \begin{tabular}{l cccc} \toprule
         \textbf{Method} & \textbf{DecoyMNIST} & \textbf{Folktables} & \textbf{MIMIC-IV} & \textbf{Pendulum}   \\ \midrule
Isotropic (L)  & $0.4963\pm0.0151$
&$0.0231\pm0.0025$
&$0.2932\pm0.0091$
&$179.6625\pm15.9685$\\ \midrule
Banana + 1st Moment & $0.0288\pm0.0045$
&$0.0111\pm0.0049$
&$0.0034\pm0.0026$
&$970.2352\pm277.8245$ \\
Banana + 1st and 2nd Moment & $0.0291\pm0.0016$
&$0.0218\pm0.003$
&$0.0857\pm0.024$
&$339.4531\pm56.9897$ \\
Banana + MMD & $0.0288\pm0.0044$
&$0.02\pm0.0048$
&$0.004\pm0.002$
&$0.0035\pm0.0055$ \\
Banana + 1st Moment (SWAG) & $0.0015\pm0.0005$
&$0.0014\pm0.0021$
&$0.0\pm0.0$
&$0.0\pm0.0$ \\ \bottomrule
    \end{tabular}
    }
    \label{tab:sample_prior_extend}
    }
\end{table*}

\subsection{Comparison against Frequentist Approaches} \label{appendix:frequentist}

While not the main focus of our paper, we also provide a comparison against standard frequentist approaches to incorporate domain knowledge. We compare against a standard supervised learning approach and a \textbf{Lagrangian}-penalized approach, where we can directly regularize with the value of $\phi$ times some hyperparameter $\lambda$, as in Eq. \eqref{equation:lagrangian}. We also consider an ensemble of such Lagrangian-penalized methods, which we refer to as \textbf{Lagrangian ensemble}. We also remark that this would be similar to the performance of posterior regularization.

\begin{table*}[h]
    \centering
    \caption{We compare Banana against frequentist analogues that incorporate domain knowledge and report the accuracy, AUROC, or $L_1$ loss and $\phi$ ($\pm$ s.e.) when averaged over 5 seeds.}
    \setlength{\tabcolsep}{4pt}
    {\renewcommand{\arraystretch}{1.1}
    \resizebox{\columnwidth}{!}{%
    \begin{tabular}{l cc cc cc} \toprule
         & \multicolumn{2}{c}{\textbf{DecoyMNIST}} & \multicolumn{2}{c}{\textbf{MIMIC-IV}} & \multicolumn{2}{c}{\textbf{Pendulum}} \\
         \cmidrule(lr){2-3} \cmidrule(lr){4-5} \cmidrule(lr){6-7} 
         Method & Accuracy ($\uparrow$) & $\phi_{\text{background}}$ &  AUROC ($\uparrow$) & $\phi_{\text{clinical}}$ & $L_1$ Loss ($\downarrow$) & $\phi_{\text{energy\_damping}}$ \\ 
         \midrule
          Supervised                     &  70.54 $\pm$ 1.64  & 0.31 $\pm$ 0.08 & 0.6987 $\pm$ 0.0011 & 0.1697 $\pm$ 0.0131 & 0.150 $\pm$ 0.049 & 0.0083 $\pm$ 0.0026    \\
         Lagrangian  & 78.98 $\pm$ 1.39 & 0.13 $\pm$ 0.01 & 0.6993 $\pm$ 0.0026 & 0.1684 $\pm$ 0.0026 & 0.186 $\pm$ 0.058 & 0.0035 $\pm$ 0.0014 \\
         Lagrangian Ens. & 81.80 $\pm$ 0.36 & 0.14 $\pm$ 0.01   & 0.6982 $\pm$ 0.0017 & 0.1623 $\pm$ 0.0027  & 0.00032 $\pm$ 0.00004 & 0.0043 $\pm$ 0.0007 \\
         \textbf{Banana}                & 78.21 $\pm$ 0.40 & 0.44 $\pm$ 0.01  & 0.6983 $\pm$ 0.0001 & 0.1619 $\pm$ 0.0005 & 0.0041 $\pm$ 0.0007 & 0.0025 $\pm$ 0.0010 \\ \bottomrule
 \end{tabular}
    }
    }
    \label{tab:accs_phi_lag}
\end{table*}



In Table \ref{tab:accs_phi_lag}, we observe it seems more effective to directly regularize with $\phi$ in making the values of $\phi$ smaller. On the MIMIC dataset, we observe that all methods are comparable. This is likely due to the domain knowledge not being particularly helpful given the amount of labeled data (2000 examples); this is supported by the observation that the supervised and lagrangian methods have similar performance. However, we again note that performing Lagrangian ensembling methods are more computationally intensive, as it requires regularizing with $\phi$ during each model training process.

\renewcommand\thetable{B\arabic{table}}
\renewcommand\thefigure{B\arabic{figure}}
\setcounter{table}{0}
\setcounter{figure}{0}
\section{Experimental Details}\label{sec:exp-details}

\paragraph{Hyperparameters} We perform hyperparameter optimization over the following hyperparameter values, selecting the best-performing method on the validation set. For all methods, we consider two-layer neural networks with a ReLU activation function and a hidden dimension size $ \in [8, 16, 32]$ for the Folktables dataset, $\in [8]$ for DecoyMNIST and Pendulum, and $\in [32, 64, 96]$ for MIMIC-IV. We also consider batch sizes in $[129, 256, 512]$ for Folktables, $[128, 256]$ for DecoyMNIST and Pendulum, and $[128, 256, 512]$ for MIMIC-IV. We remark that on DecoyMNIST the value of gradients (with respect to input data) is quite sensitive to the overall scale of the for the learnable parameters of the informative prior in Banana. 
Therefore, we use an initialization randomly sampled from $\mathcal{N}(0, 0.01)$.
We also use a $N(0, 0.01)$ initialization for Pendulum. 
On other tasks, we simply initialize the parameters with $\mathcal{N}(0, 1)$ as they are not as sensitive. For BNNs, we similarly consider a prior distribution of $\mathcal{N}(0, \sigma^2 I)$, where $\sigma^2$ is a hyperparameter tuned on the validation set. For specific methods, we use the following hyperparameters. 

\paragraph{Supervised and Lagrangian}
\begin{itemize}[itemsep=2pt]
    \vspace{-2mm}
    \item learning rate $\in [0.1, 0.01, 0.001, 0.0001]$ on Folktables Pendulum and MIMIC; learning rate $\in [0.01, 0.001, 0.0001]$ on DecoyMNIST
    \item epochs $\in [10, 20, 30]$ on Folktables, Pendulum, and MIMIC; epochs $\in [10, 15, 20]$ on DecoyMNIST
    \item $\lambda \in [1, 0.1, 0.01, 0.001]$ on Folktables, Pendulum, and MIMIC;  $\lambda \in [0.001, 0.0001]$ on DecoyMNIST
    \item weight decay $\in [0, 0.01, 0.1]$, usedin a standard $L_2$ penalization over network weights
\end{itemize}

\paragraph{BNN and Banana}  
\begin{itemize}[itemsep=2pt]
    \vspace{-2mm}
    \item number of models (posterior samples) $= 5$ on Folktables, Pendulum, and MIMIC; number of models $ = 3$ on DecoyMNIST
    \item pretraining epochs $\in [10, 20, 30]$
    \item posterior epochs $\in [30, 40, 50]$ in MIMIC; posterior epochs $\in [5, 10]$ in DecoyMNIST; posterior epochs $\in [50, 75, 100]$ on Folktables; posterior epochs $\in [10, 20, 30]$ on Pendulum
    \item prior weight $\in [10^{-7}, 10^{-8}, 10^{-9}, 10^{-11}]$ on MNIST; prior weight $\in [10^{-7}, 10^{-9}, 10^{-11}]$ on Folktables; prior weight $\in [10^{-6}, 10^{-7}, 10^{-8}]$ on MIMIC; prior weight $\in [10^{-7}, 10^{-8}, 10^{-9}]$ on Pendulum
    \item $\tau =1$ for MIMIC; $\tau \in [0.001, 0.0005, 0.0001]$ for Pendulum; $\tau \in [10, 1, 0.1]$ for DecoyMNIST; $\tau \in [0.1, 0.01, 0.001]$ for Folktables
    \item $\beta \in [0.01, 0.001, 0.0001]$ for Folktables and DecoyMNIST; $\beta \in [0.1, 0.01, 0.001]$ for MIMIC; $\beta \in [0.01, 0.001]$ for Pendulum
    \item low-rank approximation $r \in [20, 30]$ for DecoyMNIST, Folktables, and MIMIC; $r \in [10, 20]$ for Pendulum
    \item prior learning rate $\in [0.1, 0.01]$ on Folktables and Pendulum; prior learning rate $\in [0.01, 0.001, 0.0001]$ on DecoyMNIST; prior learning rate $\in [0.01, 0.001]$ on MIMIC
    \item posterior learning rate $\in [0.5, 0.1, 0.01]$ on Folktables; posterior learning rate $\in [0.01, 0.005, 0.001]$ for Pendulum; posterior learning rate $\in [0.01, 0.001, 0.0001]$ for DecoyMNIST; posterior learning rate $\in [0.5, 0.3, 0.1]$ for MIMIC
    \item $\sigma^2 \in [0.01, 0.1, 1]$ for DecoyMNIST; $\sigma^2 = 1$ for Folktables, Pendulum, and ACS
\end{itemize}

\paragraph{Compute Resources} Each experiment was run on a single GeForce 2080 Ti GPU.



\renewcommand\thetable{C\arabic{table}}
\renewcommand\thefigure{C\arabic{figure}}
\setcounter{table}{0}
\setcounter{figure}{0}
\section{Dataset Details}

\subsection{Details on Folktables}

We use the Folktables dataset for the task of determining the employment of a particular job applicant. We restrict our focus to the Alabama subset of the data from 2018. We refer readers to \citep{ding2021retiring} for more specific details about the dataset and its collection.

\subsection{Details on Pendulum Dataset}

On the Pendulum dataset \citep{seo2021controlling}, we use the configuration detailed in Table \ref{tab:pend_tab} for generating the time-series data. We refer the readers to \citet{seo2021controlling} for full details on the dataset.


\begin{table}[h]
    \caption{Constants used in the generation of the Pendulum dataset.}
    \centering
    \vspace{10pt}
    \begin{tabular}{l c}\toprule 
        \textbf{Dataset Configuration} & \textbf{Value} \\ \midrule
        String 1 Length & 1  \\
        String 2 Length & 1 \\
        Mass 1 & 1 \\
        Mass 2 & 5 \\
        Friction Coefficient 1 & 0.001 \\
        Friction Coefficient 2 & 0.001 \\ \bottomrule
    \end{tabular}
    \label{tab:pend_tab}
\end{table}

\subsection{Details on Healthcare Data}\label{appendix-healthcare}


MIMIC-IV \citep{mimic-iv} is an open-access database that consists of deidentified electronic health record data collected at the Beth Israel Deaconness Medical Center between years 2008 and 2019, covering over 400,000 distinct hospital admissions. For the intervention prediction task described in Section \ref{sec:data}, we focus on admissions that include a stay in the intensive care unit (ICU), for which various physiological measurements from bedside monitors, lab tests, etc. are readily available at higher temporal resolution. We provide details on how the study cohort was selected for the experiments, how the features and labels were extracted, and a demographics summary of the final resulting cohort. 

\paragraph{Cohort Selection.} For our study cohort, we include all ICU stays that satisfy the following criteria:
\begin{itemize}[itemsep=2pt]
    \item Adult patients: Given that physiology of young children and adolescents can differ significantly from that of adults, we only include ICU stays corresponding to adult patients between the age of 18 and 89 at the time of admission.
    \item First ICU stay: Following standard practice \citep{mimic-extract}, if a patient has multiple ICU stays recorded in the database across all hospitalizations, we only include the first ICU stay.
    \item Length of ICU stay $ \geq$ 48 hours: We only include ICU stays that lasted long enough to have a sufficient number of measurements for every stay and remove outlier cases.
\end{itemize}

We note that not all ICU stays selected by this inclusion criteria are eventually included, due to the additional filtering steps detailed in the description on feature and label extraction below. We include a summary of demographic information for the final extracted cohort in Table \ref{tab:mimic-demographics}.

\begin{table}[t]
    \caption{Summary of demographics for the final extracted cohort of ICU patients. Except for the total number of ICU patients included, we report the mean and standard deviation (in parentheses) of each demographic feature.}
    \centering
    \vspace{10pt}
    \begin{tabular}{llll}
    \toprule
                     &                     & Missing &        Overall \\
    \midrule
    Number of ICU Patients & {} &         &          13944 \\
    \midrule
    Age &                     &       0 &    64.3 (15.7) \\
    Gender & Female &       0 &    5751 (41.2) \\
                     & Male &         &    8193 (58.8) \\
    Ethnicity & Asian &       0 &      394 (2.8) \\
                     & Black &         &    1464 (10.5) \\
                     & Hispanic &         &      525 (3.8) \\
                     & Native American &         &       57 (0.4) \\
                     & Other/Unknown &         &    2451 (17.6) \\
                     & White &         &    9053 (64.9) \\
    Admission Height &                     &    3843 &   169.7 (10.5) \\
    Admission Weight &                     &       0 &    84.0 (25.1) \\
    Length of Stay &                     &       0 &  185.5 (185.4) \\
    ICU Type & Cardiac Vascular Intensive Care Unit (CVICU) &       0 &    2317 (16.6) \\
                     & Coronary Care Unit (CCU) &         &    1625 (11.7) \\
                     & Medical Intensive Care Unit (MICU) &         &    3432 (24.6) \\
                     & Medical/Surgical Intensive Care Unit (MICU/SICU) &         &    2237 (16.0) \\
                     & Neuro Intermediate &         &       44 (0.3) \\
                     & Neuro Stepdown &         &       16 (0.1) \\
                     & Neuro Surgical Intensive Care Unit (Neuro SICU) &         &      386 (2.8) \\
                     & Surgical Intensive Care Unit (SICU) &         &    1933 (13.9) \\
                     & Trauma SICU (TSICU) &         &    1954 (14.0) \\
    \bottomrule
    \end{tabular}
    \label{tab:mimic-demographics}
\end{table}

\paragraph{Feature and Label Extraction.} For all ICU stays included in the cohort, we extract the same set of features (2 static and 6 time-dependent features) used in \citet{output-bnn-2020}, listed below.
\begin{itemize}[itemsep=2pt]
    \item \texttt{Mean Arterial Pressure (MAP)}: Time-Dependent
    \item \texttt{Age at Admission}: Static
    \item \texttt{Urine Output}: Time-Dependent
    \item \texttt{Weight at Admission}: Static
    \item \texttt{Creatinine}: Time-Dependent
    \item \texttt{Lactate}: Time-Dependent
    \item \texttt{Bicarbonate}: Time-Dependent
    \item \texttt{Blood Urea Nitrogen (BUN)}: Time-Dependent
\end{itemize}
Every recorded time-dependent feature has an associated time stamp (e.g., \texttt{2180-07-23 23:50:47}), and we use the measurement time offset from the start time of the corresponding ICU stay to aggregate all measurements into hourly bins and obtain a discretized time-series representation. Within each hourly bin, if more than one measurements are available, we take the most recent measurement. For the static features, we duplicate them along all hourly bins. For example, suppose that a patient's first ICU stay lasted for 2 days. We then obtain a $48 \times 8$ time-series representation, where the rows correspond to the 48 hourly bins, and the columns correspond to the 6 time-dependent features and the 2 static features duplicated along all rows.

As in \citet{output-bnn-2020}, we consider a \textit{time-independent} binary classification task, where we treat the 8-dimensional features at each hourly bin as a separate sample and predict whether an intervention for hypotension management is necessary for the given hour. To obtain the hourly labels to predict, we extract the start and end times of all recorded vasopressor (e.g., norepinephrine, dobutamine) administrations for each ICU stay, and label each hourly bin as 1 if the vasopressor duration coincides with the hourly bin. 

Additionally, given that clinical measurements are measured at different intervals and high levels of sparsity, we filter out all rows that have missing features. Concatenating all samples together, we obtain input features $X \in \mathbb{R}^{49953 \times 8}$ and labels $Y \in \{0,1\}^{49953}$, where the labels are approximately balanced (positive: 25171 samples, negative: 23565 samples). We then take a stratified 70-15-15 split to get the training, validation, and test datasets while preserving the label proportions, and standardizing all features to zero mean and unit variance based on the training data. We also note that we take a subset of the training data when used to compare all of our approaches; we use a total of 2000 examples.

\paragraph{Thresholds Used for Defining $\phi_{\text{clinical}}$.} In adults, the normal range for lactate levels are $0.5$--$2.2$ mmol/L\footnote{\url{https://www.ucsfhealth.org/medical-tests/lactic-acid-test}} 
and bicarbonate levels are $22$--$32$ mmol/L\footnote{\url{https://myhealth.ucsd.edu/Library/Encyclopedia/167,bicarbonate}}, and we therefore define the thresholds $\tilde{x}_{\text{lactate}} = 2.2$ and $\tilde{x}_{\text{bicarbonate}} = 22$ and \textit{standardize these values according to the training data}. 




\end{document}